\documentclass[10pt,twocolumn,letterpaper]{article}

\usepackage{wacv}
\usepackage{times}
\usepackage{epsfig}
\usepackage{graphicx}
\usepackage{amsmath}
\usepackage{amssymb}
\usepackage[inline]{enumitem}
\usepackage{makecell}
\usepackage{multirow}
\usepackage{hhline}
\usepackage[accsupp]{axessibility}

\def\wacvPaperID{1188} 

\wacvfinalcopy 

\ifwacvfinal
\fi

\ifwacvfinal
\usepackage[breaklinks=true,bookmarks=false]{hyperref}
\else
\usepackage[pagebackref=true,breaklinks=true,colorlinks,bookmarks=false]{hyperref}
\usepackage{xcolor}
\fi

\ifwacvfinal
\fi

\begin{document}

    \title{Revealing Disocclusions in Temporal View Synthesis through Infilling Vector Prediction}

    \author{Vijayalakshmi Kanchana\footnotemark[1]
    \quad Nagabhushan Somraj\footnotemark[1] \quad Suraj Yadwad \quad Rajiv Soundararajan\\
    Indian Institute of Science, Bengaluru, India\\
    {\tt\small \{vijayalaksh1, nagabhushans, surajyadwad, rajivs\}@iisc.ac.in}
    }
    \maketitle
    
    
    
\ifwacvfinal{
    \renewcommand{\thefootnote}{\fnsymbol{footnote}}
    \footnotetext{Project Page: \url{https://nagabhushansn95.github.io/publications/2021/ivp.html}}
    \setcounter{footnote}{1}
    \footnotetext{The authors contributed equally to this work.}
} \fi
    
    \begin{abstract}
        We consider the problem of temporal view synthesis, where the goal is to predict a future video frame from the past frames using knowledge of the depth and relative camera motion.
        In contrast to revealing the disoccluded regions through intensity based infilling, we study the idea of an infilling vector to infill by pointing to a non-disoccluded region in the synthesized view.
        To exploit the structure of disocclusions created by camera motion during their infilling, we rely on two important cues, temporal correlation of infilling directions and depth. 
        We design a learning framework to predict the infilling vector by computing a temporal prior that reflects past infilling directions and a normalized depth map as input to the network. 
        We conduct extensive experiments on a large scale dataset we build for evaluating temporal view synthesis in addition to the SceneNet RGB-D dataset.
        Our experiments demonstrate that our infilling vector prediction approach achieves superior quantitative and qualitative infilling performance compared to other approaches in literature.
    \end{abstract}
    \section{Introduction}\label{sec:introduction}
    Suppose a user is exploring a virtual environment on a head mounted display device.
    Given the past view of a rendered video frame and the most recent head position, we ask if the updated view can be generated directly without  graphically rendering the frame again.
    We refer to the above problem as egomotion aware temporal view synthesis.
    The problem has applications in frame rate upsampling for virtual reality (VR) and gaming applications in low compute devices through positional time warp \cite{Asynchro98:online}, asynchronous reprojection \cite{Asynchro19:online} or interleaved reprojection \cite{valve-interleaved}.

\begin{figure*}
        \centering
        \includegraphics[width=0.9\linewidth]{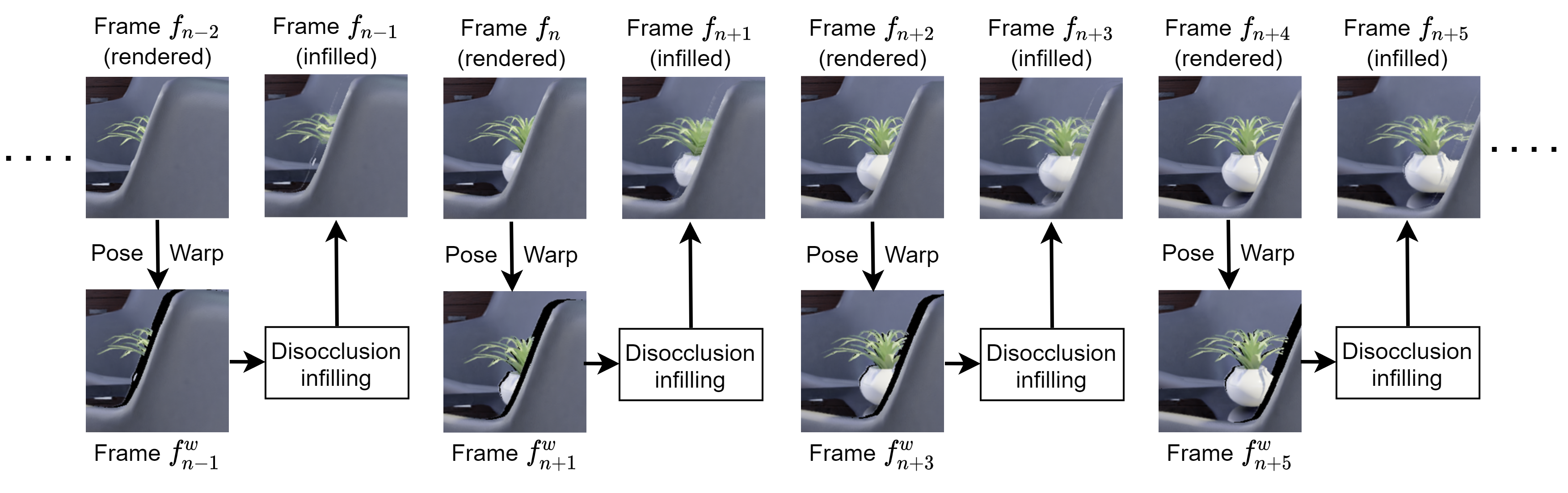}
        \caption{\textbf{Temporal view synthesis for frame rate upsampling}. Graphically rendered frame $f_n$ is pose warped using relative pose and depth. The disocclusions in the warped frame are infilled to obtain the filled frame $f_{n+1}$. Frame $f_{n+2}$ is now again graphically rendered and $f_{n+3}$ is predicted and so on.}
        \label{fig:Top-summary-fig}
    \end{figure*}

    Given the head pose (or 6D camera pose) and the past frame along with its depth map, we can reconstruct the next frame by warping the past frame through a transformation matrix.
    During this warping, the disocclusion of the background leads to missing pixels in the synthesized view as shown in Figure \ref{fig:Top-summary-fig}.
    While newer regions on the boundary can also emerge during temporal view synthesis, VR applications can solve this by starting with a larger field of view and cropping out the desired portion. An important distinction between novel view synthesis \cite{zhou2016view} and temporal view synthesis is the high frame rate of the videos. This primarily introduces disocclusions of the relative background regions as opposed to those of the foreground object \cite{zhou2016view}.
    We particularly focus on the problem of revealing disocclusions of the background during temporal view synthesis.

    One solution for infilling disocclusions is based on mesh based interpolation methods such as the reference view synthesizer (RVS) provided by the moving pictures experts group (MPEG) \cite{fachada2018depth, mpeg2018rvs} or splatting based methods \cite{botsch2005high, zwicker2001surface}.
    The disoccluded or unknown regions in such solutions tend to get stretched, leading to geometric distortions and temporal inconsistencies in the synthesized video.
    Novel view synthesis methods \cite{penner2017soft, wiles2020synsin, zhou2018stereo} typically assume that the depth maps in the past frames are not available and are designed to overcome this challenge.
    The novel view synthesis problem also does not consider the temporal dependencies that can be exploited in temporal view synthesis.
    
    \begin{figure}
        \centering
        \includegraphics[width=0.85\linewidth]{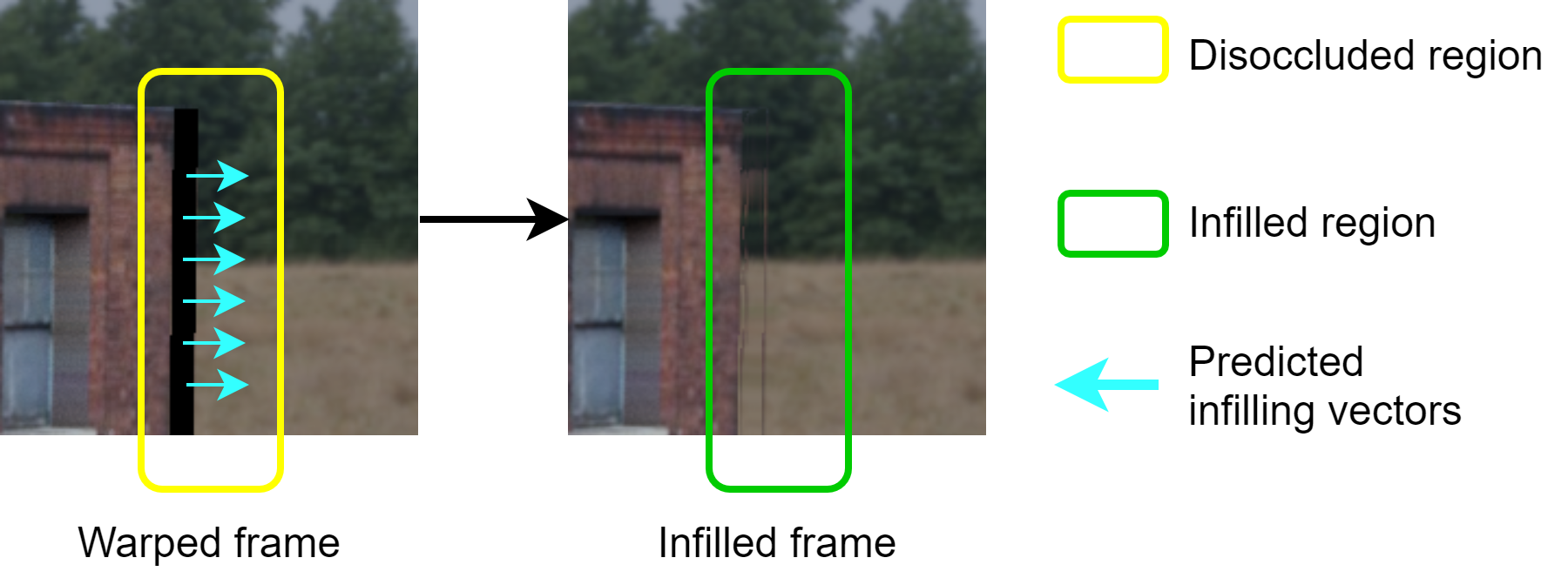}
        \caption{\textbf{Infilling Vector}. The disocclusions in the warped frame can be filled using the infilling vector from the known regions to the unknown regions.}
        \label{fig:infilling-vector-idea}
    \end{figure}

    Alternatively, one could employ generic image inpainting to predict the missing pixels in the warped frame \cite{Nazeri_2019_ICCV, shih20203d, zeng2020high}.
     The causal constraint in temporal view synthesis limits the application of video inpainting methods \cite{kim2019deep, xu2019deep}.
    Further, both image and video inpainting methods are not designed to exploit the extra information available in camera motion and depth for VR applications. 
    While several depth image based rendering approaches (DIBR) \cite{daribo2011novel, luo2019disocclusion} exploit depth information for infilling the disocclusions, there is little work on learning based methods.

    Our approach is motivated by the intuition that disoccluded pixels can be infilled by copying intensities from the spatial neighborhood.
    We introduce the idea of an infilling vector for revealing disocclusions.
    An infilling vector points to the known regions of the warped frame from where intensities in the unknown regions can be read off as shown in Figure \ref{fig:infilling-vector-idea}.
    While deep video inpainting \cite{kim2019deep} uses flow vectors to infill missing regions visible in the past frames, it does not sufficiently exploit the structure of missing pixels created due to the disocclusions caused by camera motion.
    
    Our main contribution is in exploiting the structure of the disocclusions created by camera motion and the available scene depth to guide the infilling through a deep network. The continuity of camera motion results in a similarity in the infilling directions across frames. So we first estimate a fictitious infilling vector in a rendered frame warped from a previously rendered frame.
    We then use this estimated infilling vector to temporally guide the prediction of the desired infilling. 
    We also observe that the disoccluded regions can be reconstructed by pointing to the relative background regions in their neighborhood. 
    We encourage such infilling by obtaining a normalized scene depth in a normalized range that guides the infilling vector prediction.  

    While several datasets exist for studying the novel view synthesis problem, the videos have low spatial resolutions.
    Thus, we create a new Indian Institute of Science Virtual Environment Exploration Dataset, IISc-VEED, of graphically rendered videos at a spatial resolution of 1920x1080 and a temporal resolution of 30 frames per second.
    Our dataset consists of 800 video clips of 12 frames each.
    We carefully designed this dataset to particularly reveal the challenges of infilling disocclusions.
    We conduct detailed experiments on our dataset and the SceneNet RGB-D datasets to show that our model outperforms other competing methods.
    We also propose a new measure to evaluate the temporal consistency of the infilled regions, and show that our model generates temporally consistent infilled regions.

    We summarize our main contributions as follows:
    \begin{enumerate}
        \item We design a novel deep infilling vector prediction algorithm that exploits the temporal correlation of infilling directions in successive frames. 
        \item We also guide the infilling vectors to point to the relative background by computing a normalized depth map as input to our infilling vector prediction method. 
        \item We construct a new dataset, IISc-VEED, to evaluate disocclusion infilling in temporal view synthesis.
        \item We benchmark several view synthesis methods for frame-rate upsampling on our dataset and the SceneNet RGB-D dataset~\cite{mccormac2017scenenet}. We show that our model outperforms all other competing approaches while yielding more temporally consistent predictions.
    \end{enumerate}

    \section{Related Work}\label{sec:related-work}
    We now discuss several strands of work that are related to our problem as follows.  

    \textbf{Novel View Synthesis.} Novel view synthesis from single or multiple given views is relevant to our problem by treating the past frame(s) as the given view(s) and the future frame to be predicted as the novel view.
    Since the depth is often assumed not known, these methods attempt to estimate the depth through probabilistic depth volumes \cite{choi2019extreme}, multi-plane images (MPI) \cite{flynn2019deepview, srinivasan2019pushing, zhou2018stereo} and soft 3D reconstructions \cite{penner2017soft}.
    Disocclusions in novel views have been carefully handled during construction of MPI planes by using flow vectors that point to visible pixels \cite{srinivasan2019pushing}.
    Instead of estimating depth, view synthesis methods also deal with the use of incomplete or noisy depth information \cite{novotny2019perspectivenet}.
    Several view synthesis methods for generating views of individual objects from single views exist in literature \cite{park2017transformation, sun2018multi, zhou2016view}.
    View synthesis of more generic scenes from single views has also been designed \cite{chen2019monocular, wiles2020synsin}.
    Depth is first estimated and used to warp the views in either the pixel domain \cite{chen2019monocular} or the feature space \cite{wiles2020synsin}.

    \textbf{Image and Video Inpainting.} Off-the-shelf image inpainting algorithms can be applied on the warped frames containing holes.
    Image inpainting is a well studied topic with rich literature ranging from classical methods \cite{criminisi2004region, guillemot2013image} to modern deep learning based methods.
    Deep inpainting methods based on adversarial approaches have achieved state of the art performance \cite{Nazeri_2019_ICCV, yu2018generative}.
    While video inpainting and completion methods \cite{gao2020flow, kim2019deep, liao2020dvi, xu2019deep} are also related to the problem of hole filling in warped frames, these methods do not adhere to the causal constraint which exists in temporal view synthesis. Further, they do not exploit the structure of holes created due to disocclusions caused by camera motion. 

    \textbf{Depth Image based Rendering.}  Techniques that enable free viewpoint 3D video through DIBR face similar challenges to the problem of temporal view synthesis.
    Disocclusion infilling approaches in DIBR
    can be classified as spatial infilling approaches that are depth aware \cite{ahn2013novel, cho2017hole, daribo2011novel} and spatio-temporal approaches that focus on reconstructing the background before warping and infilling \cite{luo2019disocclusion}.
    Nevertheless, all the above methods are not learning based and relevance of deep learning has not been explored much.
    While deep networks are used to correct depth errors in DIBR \cite{cai2020hole}, the problem of disocclusion infilling through deep networks has not been studied to the best of our knowledge.
    
    \textbf{Video Prediction.} Although temporal view synthesis is close to video prediction \cite{mathieu2016deep,oprea2020review}, our goal is disocclusion infilling due to effective use of camera motion, depth and scene geometry. In contrast, video prediction methods aim to predict all the pixels in successive frames. They do not make effective use of the camera motion and geometry. 
    
    \section{Problem Statement}
    We now describe the temporal view synthesis problem and the notation.
    We specifically focus on the scenario where motion in the video occurs on account of egomotion (due to a moving camera or head movement).
    We assume that the scene is static and the objects are stationary.
    Let frame $f_n$ be given, along with its depth map $d_n$, and relative $4\times4$ transformation matrix $T_n$ from  $f_n$ to $f_{n+1}$.
    Our goal is to predict frame $f_{n+1}$.
    To predict $f_{n+1}$, we leverage projective geometry to first warp homogeneous pixel coordinates in $f_n$ to the view of $f_{n+1}$, by using the transformation matrix and the depth map as
    \begin{equation}
        p^w_{n+1} \sim K\mathcal{G}\{T_n\mathcal{F}\{d_n(p_n)K^{-1}p_n\}\}, \label{eqn:warping}
    \end{equation}
    where $K$ is the $3\times3$ camera intrinsic matrix, $p_n$ is a 2D image location expressed in homogeneous form and $p^w_{n+1}$ is the warped image location in $f_{n+1}$.
    $\mathcal{F}$ and $\mathcal{G}$ denote the conversion from $3\times1$ to $4\times1$ vectors and vice versa, respectively.
    Using the above, we reconstruct the pixel intensities in $f_{n+1}$, corresponding to the regions which can be mapped to $f_n$ by interpolating from the transformed coordinates in the immediate grid neighborhood.
    We refer to these regions as `known' regions.

    Warping cannot provide intensities for the locations occluded in $f_n$ which get disoccluded in $f_{n+1}$.
    This leads to the creation of holes or disoccluded regions in the warped frame as indicated by black regions in Figure~\ref{fig:infilling-vector-idea}.
    While warping cannot also predict intensities of pixels corresponding to the new regions that emerge in the next frame, one can solve this problem by rendering the frames at a larger field of view and then cropping the desired region during display.
    Hence, our primary focus is on infilling the disoccluded regions in frames reconstructed by warping.
    Thus, the temporal view synthesis problem involves taking as input $f^w_{n+1}$ and generating the infilled frame $f^i_{n+1}$.
    We note that warping can introduce small, imperceptible errors in the known regions due to imperfect interpolation.
    Further, we assume that the frame rates are large enough that minimize the effect of illumination changes.
    Hence, we focus mainly on infilling the disoccluded regions.
    
    \textbf{Application to frame rate upsampling.} We study the application of temporal view synthesis to frame rate upsampling or interleaved reprojection.
    Suppose the graphics engine can render alternate frames $f_{n-4}$, $f_{n-2}$, $f_n$ and so on, temporal view synthesis can be used to predict $f_{n-3}$, $f_{n-1}$, $f_{n+1}$ and so on, from their immediate past rendered frames.
    This enables doubling of frame rate in a causal manner.

    \section{Method}\label{sec:method}
    We first describe our warping to reconstruct the known regions of frame $f_{n+1}$ and then describe our approach of infilling the holes created due to disocclusions.
    Our key contribution is in exploiting the structure of the disocclusions created due to egomotion in the video to achieve superior infilling performance. In particular, we utilize the temporal correlation of the infilling across frames and the scene depth to effectively infill the disocclusions caused by egomotion.
    We further improve the prediction by detecting and discarding any erroneous predictions and iteratively infill the remaining holes.

    We employ projective geometry based warping to reconstruct known regions of the next frame $f_{n+1}$ as in Equation \ref{eqn:warping}. 
    After obtaining the corresponding locations in frame $f_{n+1}$, we use inverse bilinear interpolation \cite{wang2018occlusion} to obtain intensities at the grid locations. In particular, for every grid location, we take the neighbors within a grid on all sides and combine them using inverse bilinear interpolation. During this interpolation, we also weigh the neighbors inversely proportional to the depth of the warped locations.  
    This can also be considered as a simpler splatting method~\cite{zwicker2001surface}.
    
    \subsection{Infilling Vector Prediction (IVP)}\label{subsec:ivp}
    We observe that the disoccluded regions are best infilled based on the known regions in their neighborhood.
    Thus, we resort to predicting an `infilling vector' for every pixel in the disoccluded region, which points to a pixel location in the known region, as shown in Figure~\ref{fig:infilling-vector-idea}.
    The infilling vector can then be used to predict the intensity at the desired disoccluded location.
    If $(a_{n+1}(x,y), b_{n+1}(x,y))$ are the horizontal and vertical components of the infilling vectors, then
    \begin{equation}
        f^i_{n+1}(x,y) = f^w_{n+1}(x + a_{n+1}(x,y), y + b_{n+1}(x,y)). \label{eqn:spatial-transformer}
    \end{equation}
    Our approach of predicting infilling vectors instead of the intensities in the disoccluded regions may not suffer from distortions such as blur and color gradations that occur when the intensity is directly predicted.
    Further, disoccluded regions often correspond to background pixels and our infilling vector prediction approach can indicate whether the infilled pixel points to a foreground or background location.
    Such information can be potentially used to constrain and improve the estimates.
    
    Our infilling vectors which point to the known regions within the frame are quite different from flow based models used to read off matching points from neighboring frames in deep video inpainting \cite{kim2019deep}. In the following, we first describe how the infilling vector prediction can benefit from temporal guidance and then discuss depth guidance. We show the overall architecture for our approach in Figure \ref{fig:model-architecture}.
    
    \subsection{Temporal Guidance}\label{subsec:temporal-guidance}
    Since egomotion in a video tends to be correlated across time, the shape of the disoccluded regions and the infilling direction from the background also tend to be similar across successive frames.
    In order to exploit the correlation of infilling directions across time, we input fictitious infilling vectors in previously rendered frames to predict the desired infilling vector in $f_{n+1}^{w}$ as follows.
    Since we focus on frame rate upsampling, the ground truth is available for every alternate frame.
    Using the transformation matrix, we warp $f_{n-2}$ to the view of $f_n$, denoted as $f^w_n$.
    We then use the ground truth $f_n$, to estimate the infilling vectors for the disoccluded regions in $f^w_n$.
    These fictitious infilling vectors thus estimated serve as a prior to predict the desired infilling vectors in $f^w_{n+1}$.
    We provide further details of each of the above steps in the following.

    \textbf{Estimation of infilling vectors in previous frame.}  For every disoccluded pixel at $(x,y)$ in the warped frame $f^w_n$, we search for the nearest neighbor in the known regions of the frame, in the four cardinal directions.
    We compare the intensities at these four locations with the true intensity at $(x,y)$, available from the ground truth frame $f_n$.
    We pick the neighbor that has the least mean squared error (MSE) with the ground truth value, and set the infilling vector at $(x,y)$ to point to this optimal neighbor's location.
    To avoid noisy estimates, we consider a small patch around the pixels, instead of individual pixels, and compute the MSE between the patches.
    For our experiments, we set the patch size to $3 \times 3$.
    These estimated infilling vectors $(\alpha_n,\beta_n)$ act as a strong prior to predict the infilling vectors for $f_{n+1}$.
    
    \textbf{Learning infilling vectors with temporal guidance.} We exploit temporal similarity by feeding the estimated infilling vectors in $f^w_n$ obtained as above to a deep network.
    However, the disoccluded regions in $f^w_n$ and $f^w_{n+1}$ are spatially displaced.
    We observe that the network can better utilize the estimated infilling vectors if they are closer to the disoccluded region locations.
    We achieve this by warping the infilling vectors $(\alpha_n,\beta_n)$ to obtain $(\alpha_{n+1}^w,\beta_{n+1}^w)$ in a manner similar to the warping of intensities.
    This brings the estimated infilling vectors closer to the disoccluded regions in $f_{n+1}^w$.
    Since the estimated infilling vectors $f_n^w$ may be noisy, we smoothen the warped vectors using an averaging low-pass filter.
    We adopt a U-Net architecture to learn the infilling vectors $(a_{n+1},b_{n+1})$ from $(\alpha_{n+1}^w,\beta_{n+1}^w)$.
    
    \subsection{Depth Guidance}
    \begin{figure*}
        \centering
        \includegraphics[width=1.0\linewidth]{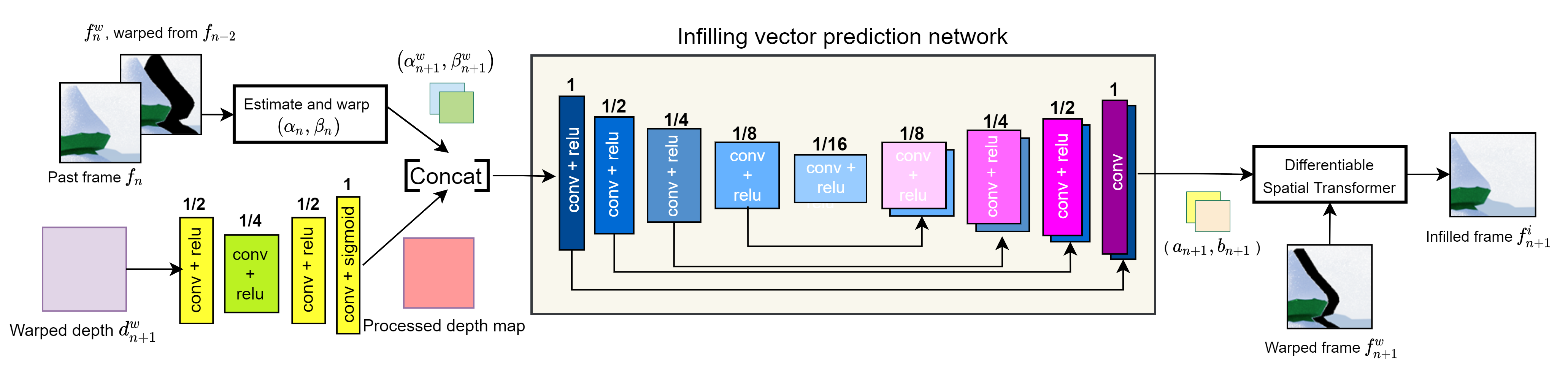}
        \caption{\textbf{Architectural pipeline of our model.} The infilling vectors are predicted by U-Net, which takes in $(\alpha_{n+1}^w,\beta_{n+1}^w)$ and processed depth map as a 3-channel input and outputs the dense 2-channel infilling vector field $(a_{n+1},b_{n+1})$. The temporal prior $(\alpha_{n+1}^w,\beta_{n+1}^w)$ is estimated using $f_n$ and $f_n^w$ warped from $f_{n-2}$.}
        \label{fig:model-architecture}
    \end{figure*}
    Since the temporal prior may not always be relevant when the direction of egomotion changes or when the videos have low frame rates, we also explore the use of depth to guide the infilling. 
    Since infilling is often performed by copying pixels from the relative background, we believe that the scene depth can provide important cues on where the infilling vector needs to point to. 
    Further, when there are multiple backgrounds at different depths, the relative depths can be carefully analyzed in conjunction with the temporal prior to guide the infilling vector prediction. 
    
    Since the actual depth of the scene can vary across different scenes, we process the scene depth through a few convolutional layers to obtain a map that lies within 0 and 1. 
    This helps provide a normalized depth range that can be better exploited along with the temporal prior to predict the infilling vectors. 
    Note that the ground truth depth map, $d_{n+1}$, is not available since that is the frame we would like to predict. Thus, we warp previous frame depth $d_n$ similar to warping of frame $f_n$ and feed the warped depth $d^w_{n+1}$ to the network.
    Although the warped depth will have holes in the disoccluded regions, the depth in the known regions is available and infilling vectors will only point to these regions. 
    
    \subsection{Overall Architecture}\label{subsec:overall-architecture}
    We show the complete architecture of our model  in Figure~\ref{fig:model-architecture}. We input the warped infilling vectors $(\alpha_{n+1}^w,\beta_{n+1}^w)$ and the processed depth to a U-Net model that outputs the infilling vectors $(a_{n+1},b_{n+1})$ at all locations. Our U-Net model consists of 4 sub-sampling layers and skip connections. The output infilling vectors at the disoccluded locations are then used to predict the intensities there using Equation (\ref{eqn:spatial-transformer}).
    The error in the intensity prediction in the disoccluded regions is used to train the U-Net.
    In particular, we use a weighted combination of mean squared error (MSE) loss, structural similarity (SSIM) and smoothness constraint
    \begin{equation}
        \label{eqn:loss}
        L = \lambda_1 L_{\textrm{MSE}} + \lambda_2 L_{\textrm{SSIM}} + \lambda_3 L_{\textrm{smoothness}},
    \end{equation}
    $L_{\textrm{MSE}}$ is given by
    \begin{equation}
        L_{\textrm{MSE}} = \sum_{x}\sum_{y}m(x,y)\left [f_{n+1}(x,y)-f^i_{n+1}(x,y)\right]^2,
    \end{equation}
    where $m(x,y) = 1$, if $(x,y)$ belongs to a disoccluded region, and $0$, otherwise.
    The SSIM loss term $L_{\textrm{SSIM}}$ with the luminance and contrast terms \cite{wang2004image} is also evaluated as above with the mask. 
    We obtain a smoothness loss $L_{\textrm{smoothness}}$ on infilling vectors in the disoccluded region alone, weighted by gradient of the true image, similar to optical flow estimation~\cite{wang2018occlusion}.
    Note that gradient propagation with respect to the above loss function requires a differentiable spatial transformer module~\cite{NIPS2015_33ceb07b} as popularly used in unsupervised optical flow algorithms \cite{jason2016back, liu2019selflow, wang2018occlusion}.
    At test time, we use the infilling vectors at the disoccluded locations to predict the intensities at those locations.
    
    \subsection{Iterative Infilling}\label{subsec:iterative-infilling}
    We observe that our model can sometimes output infilling vectors that incorrectly point to foreground objects.
    Since the discoccluded regions come from the relative background, we adopt an iterative infilling method to correct this.
    In particular, we mark all infilling vectors that point to the foreground as incorrect, by setting them to zero, and do not infill in these locations.
    We utilize the warped depth $d^w_{n+1}$, which provides depth values at all known regions, to determine if an infilling vector is pointing to a foreground object.
    We now iteratively process the infilled frame where infilling has only been performed at locations where infilling vectors point to the background pixels.
    Thus, the size of the disoccluded region reduces with each iteration and in every subsequent iteration, infilling vectors predicted in the remaining disoccluded region are used for infilling.
    We perform this iterative infilling $P$ times only at test time.
    In the last iteration, we do not remove any infilling vectors. We extend any infilling vectors pointing to the disoccluded region until it points to a known region.  

    \section{Experiments}\label{sec:experiments}
    \begin{figure*}
        \centering
        \includegraphics[width=0.9\linewidth]{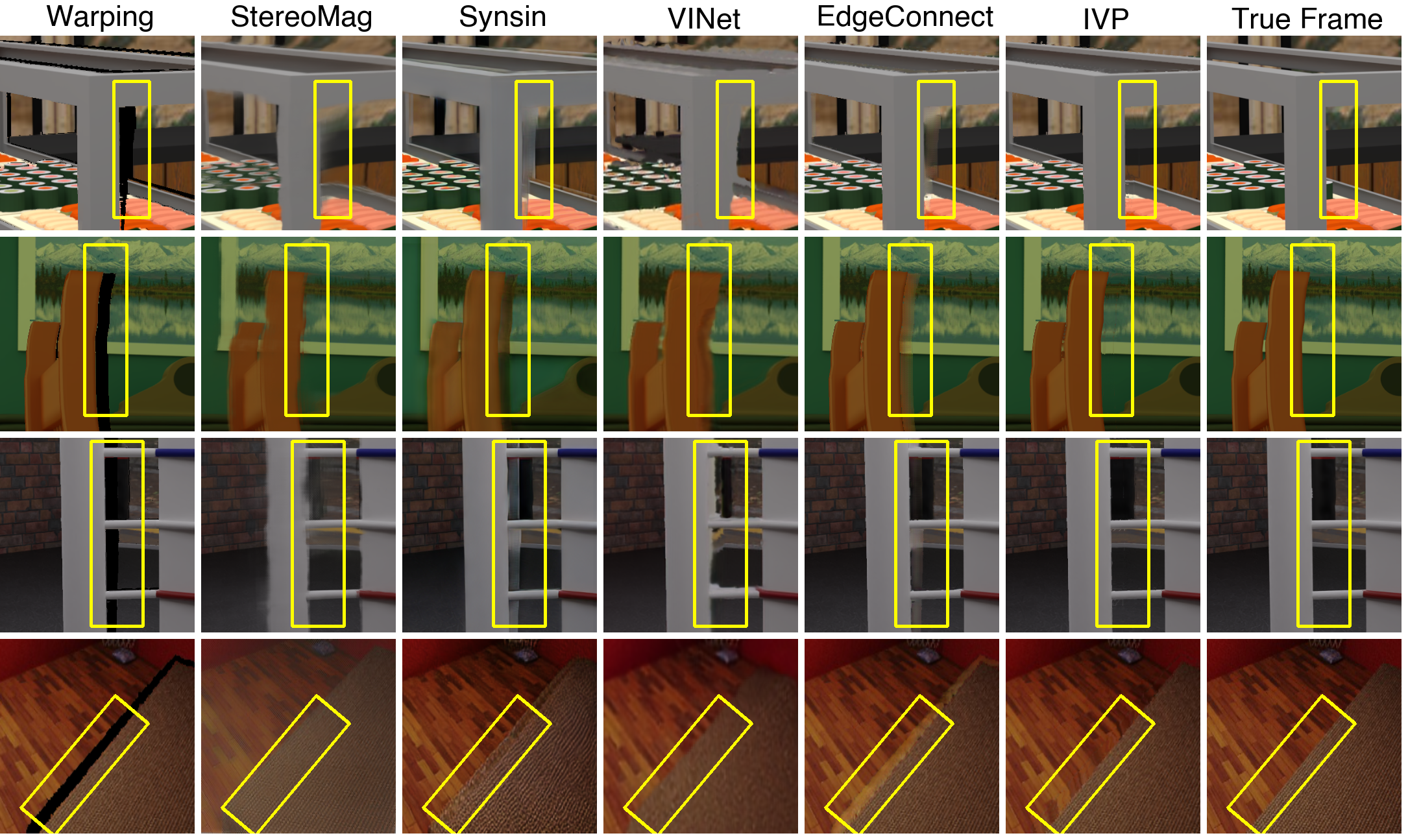}
        \caption[l]{\textbf{Comparison of temporally synthesized views by different models.}
        The first column shows warped frames, with disoccluded regions in black and the last column shows the ground truth.
        While StereoMag blurs objects in the predicted frames, SynSin and EdgeConnect predictions contain mixture of foreground and background object intensities in the disoccluded regions.
        VINet incorrectly infills the disoccluded regions with foreground objects.
        Finally, our model produces sharp and accurate results with minimal artifacts.}
        \label{fig:results-ours-bm}
    \end{figure*}
    
    \begin{table*}
        \small
        \centering
        \setlength\tabcolsep{3pt}
        \begin{tabular}{l|ccccc|ccc}
            \hline
            \multirow{2}{*}{} & \multicolumn{5}{c|}{IISc-VEED} & \multicolumn{3}{c}{SceneNet RGB-D}\\
            Model & MSE $\downarrow$ & \makecell{D-MSE} $\downarrow$ & SSIM $\uparrow$ & \makecell{D-SSIM} $\uparrow$ & \makecell{TC} $\downarrow$ & \makecell{D-MSE} $\downarrow$ & \makecell{D-SSIM} $\uparrow$ & \makecell{TC} $\downarrow$ \\
            \hline
            Copy Last Frame & 862 & 5326 & 0.7058 & 0.3296 & 1312 & 4139 & 0.2555 & 1152 \\
            RVS Warping~\cite{mpeg2018rvs} & 67 & 1192 & 0.9183 & 0.5899 & 294 & 2897 & 0.3999 & 710 \\
            \hline
            Warping + RFR~\cite{Li_2020_CVPR} & 52 & 927 & \underline{0.9251} & \underline{0.7213} &  223 & 2099 & 0.5058 & 413\\
            Warping + EC~\cite{Nazeri_2019_ICCV} & \underline{50} & 822 & 0.9238 & 0.6318 & \underline{197} & \underline{1327} & \underline{0.5222} & \underline{318} \\
            Warping + Cho \etal~\cite{cho2017hole} & 70 & 2382 & 0.9217 & 0.4910 & 570 & 2777 & 0.4414 &  1004\\
            Warping + VINet.~\cite{kim2019deep} & 72 & 2568 & 0.9209 & 0.4895 & 621 & 3872 & 0.3414 &  908\\
            \hline
            StereoMag~\cite{zhou2018stereo} & 208 & \underline{792} & 0.856 & 0.6607 & 225 & 2734 & 0.3365 &  1024\\
            SynSin~\cite{wiles2020synsin} & 127 & 1169 & 0.8273 & 0.5786 & 296 & 1707 & 0.4465 & 440\\
            3D Photography~\cite{shih20203d} & 79 & 867 & 0.9146 & 0.6872 & 219 & 2605 & 0.4029 & 731\\
            \hline
            Our Model (IVP) & \textbf{47} & \textbf{442} & \textbf{0.9262} & \textbf{0.7729} & \textbf{126} & \textbf{874} & \textbf{0.6230} & \textbf{240} \\
            \hline
        \end{tabular}
        \caption{Numerical comparisons on IISc-VEED and SceneNet RGB-D}
        \label{tab:qa-ours-short}
    \end{table*}
    \begin{figure*}
        \centering
        \includegraphics[width=0.9\linewidth]{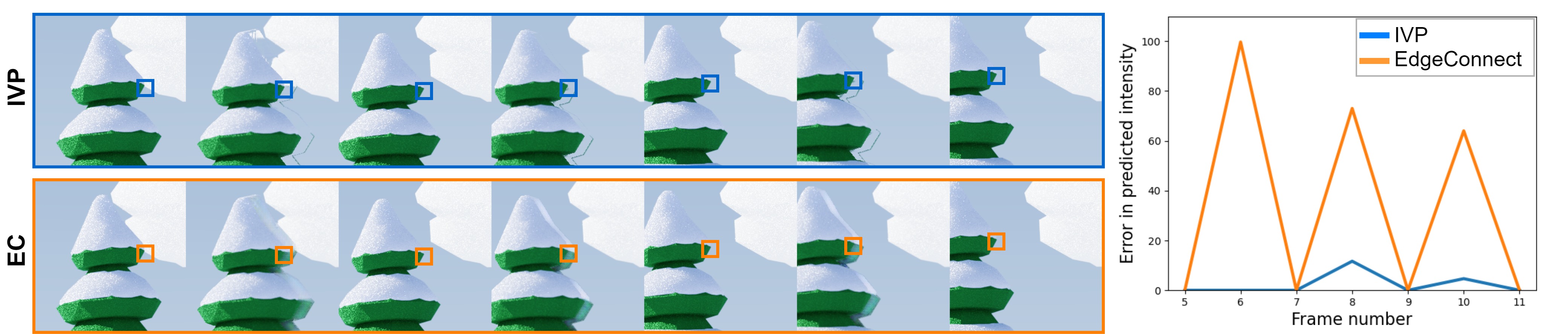}
        \caption{\textbf{Temporal consistency for the frame rate upsampling application.} The figure shows tracking of a series of pixels in the disoccluded regions, where the error in predicted intensity is measured across seven frames, of alternate ground truth and predicted frames. 
        Our model exhibits less error for the predicted frames, thereby proving that they are more temporally consistent with the ground truth frames than the other methods.
        }
        \label{fig:dev}
    \end{figure*}
    \subsection{Dataset}\label{sec:dataset-generation}
    Since our work focuses on infilling the disoccluded regions, we need a dataset that will present interesting challenges in this scenario.
    Existing datasets for view synthesis are either only available at low spatial resolutions, or do not sufficiently present the challenges in infilling disocclusions or are not appropriate for the application of temporal view synthesis in frame rate upsampling. Thus, we develop a new dataset for the temporal view synthesis problem and its application to frame rate upsampling.

    \textbf{IISc-VEED.} Our dataset consists of 800 video clips of 12 frames each at a spatial resolution of $1920 \times 1080$.
    We generate the videos using Blender by taking 200 different indoor and outdoor scene environments from \cite{bswap} and \cite{turbo} such as classroom, bedroom, kitchen, living room, cityscape, lake, seaside, windmill farm, mountains and several more.
    In each scene, we adapt the camera trajectories available in existing public datasets~\cite{handaICRA2014, sturm12iros, 7487210} and place them in the given environment to generate the video.
    The camera trajectories were chosen to introduce a moderate amount of disocclusions in the warped frames as observed for typical frame rates.
    This is repeated for four different initial viewpoints to obtain four videos per scene leading to a total of 800 videos.
    Out of the 200 scenes, we select 135 for training and the remaining 65 for testing.

    \textbf{SceneNet RGB-D Dataset.} We also evaluate and compare different methods on the SceneNet RGB-D dataset~\cite{mccormac2017scenenet}.
    This dataset consists of smaller resolution videos at $320\times240$ with larger motion between successive frames due to low frame rates.
    As a result, large parts of successive frames are dominated by new regions that enter the scene.
    Thus, we evaluate temporal view synthesis only in the disoccluded regions.
    We select 1000 videos of 12 frames from this dataset for training and 282 videos with 12 frames for testing.
    We observe that the depth information was potentially incorrect in several scenes leading to incorrect warping.
    Thus, we manually filtered 282 test videos which do not suffer from such artifacts.

    \textbf{Real vs. Synthetic Datasets.}: The problem of infilling disocclusions in warped frames is motivated by applications in graphical rendering of video frames on low compute devices. Since graphical rendering arises only in the context of synthetic scenes as opposed to real world scenes, we conduct our experiments only on synthetic scenes. Thus the depth is available. We find that infilling on synthetic scenes is still challenging and we achieve superior infilling when compared to other methods.

    \textbf{Experimental Setup.}
    We evaluate the temporal view synthesis methods for frame rate upsampling on both datasets, by assuming that $(f_{n-2},f_{n})$ are available while predicting $f_{n+1}$.
    During testing, for a video sequence of 12 frames, taking into account the need for past frames (in our model) or multiple views (in other models) and the future frames needed for evaluating temporal consistency,  we evaluate the prediction of $(f_6,f_8,f_{10})$.

    While training our model and ablations on IISc-VEED, we randomly crop out a $256 \times 256$ patch and train on patches.
    For training on SceneNet RGB-D dataset, we use the full frame, which is of resolution $320 \times 240$.
    We set the loss weights as $\lambda_1 = 1, \lambda_2 = 1, \lambda_3 = 0.001$ and train for 50 epochs.
    Additionally, at test time, we infill the frames iteratively with $P=3$, to allow for infilling correction.

    \subsection{Benchmarks for Comparison}\label{subsec:benchmarks}
    We compare our method with two sets of approaches for temporal view synthesis.
    The first set consists of novel view synthesis methods, such as Synsin~\cite{wiles2020synsin}, Stereo Magnification (StereoMag)~\cite{zhou2018stereo} and 3D Photography (3DP)~\cite{shih20203d}, which can be directly applied for temporal view synthesis given the past frame(s) and the relative camera pose and depth.
    While Synsin and 3DP use a single past frame ($f_n$) and ground truth depth ($d_n$), StereoMag uses two past frames ($f_{n-2}$ and $f_n$).
    While pre-trained models for all three methods were available, we also fine tuned them on the respective datasets if training code is available and present the best results among them.

    Since disocclusions in warped frames can be treated as missing regions, we compare by applying image and video inpainting algorithms such as Recurrent Feature Reasoning (RFR)~\cite{Li_2020_CVPR}, EdgeConnect (EC)~\cite{Nazeri_2019_ICCV}, DIBR based inpainting Cho \etal~\cite{cho2017hole} and a deep video inpainting (VINet)~\cite{kim2019deep} model for temporal view synthesis.
    We modify VINet to use only past frames and do not provide any future frames.
    We also compare with two simple baselines, one which simply copies the previous frame and RVS warping ~\cite{mpeg2018rvs,mpeg2018rvscode}.
    We do not include comparisons with video prediction algorithms since they do not use the camera motion and depth and their performance is much poorer. 

    \subsection{Performance Measures}
    We evaluate the temporally synthesized frames using mean squared error (MSE) and the structural similarity (SSIM)~\cite{wang2004image} index.
    Since the disoccluded regions account for a very small fraction of the entire frame, we also compute MSE and SSIM in these disoccluded regions only as Disoccluded-MSE (D-MSE) and Disoccluded-SSIM (D-SSIM), respectively.
    Since ground truth frames appear alternately during frame rate upsampling, the predictions in the disoccluded regions surrounding foreground objects can lead to flickering distortions.
    
    To measure the temporal consistency of infilling, as shown in Figure~\ref{fig:dev}, we first determine a sequence of pixels in successive frames at a given displacement from a location in the foreground object such that all these pixels are in the disoccluded regions in the respective frames.
    We then measure the variance of the error along this sequence as a measure of temporal consistency (TC).
    We compute this variance for several choices of the displacements and locations in the foreground object and average them.

    \subsection{Performance Comparison and Analysis}\label{subsubsec:quantitative-comparisons}
    Table~\ref{tab:qa-ours-short} shows the performance of various models with respect to the different QA measures.
    We see from the table that our model achieves superior performance when compared to inpainting and view synthesis methods, which may be attributed to the effective use of camera and depth.
    Figure~\ref{fig:results-ours-bm} shows the predictions by various state of the art models and our proposed model on test scenes from both IISc-VEED and SceneNet dataset.
    Clearly, the quality of the infilled region with our algorithm is more accurate compared to the other works considered. Since the actual video frame will appear subsequently, we do not aim to hallucinate new objects that will appear in the disoccluded regions. 
    We also note that errors in the smooth regions are more perceivable than the errors in the textured regions due to contrast masking \cite{legge1980contrast}.
    
    \begin{figure}
        \centering
        \includegraphics[width=\linewidth]{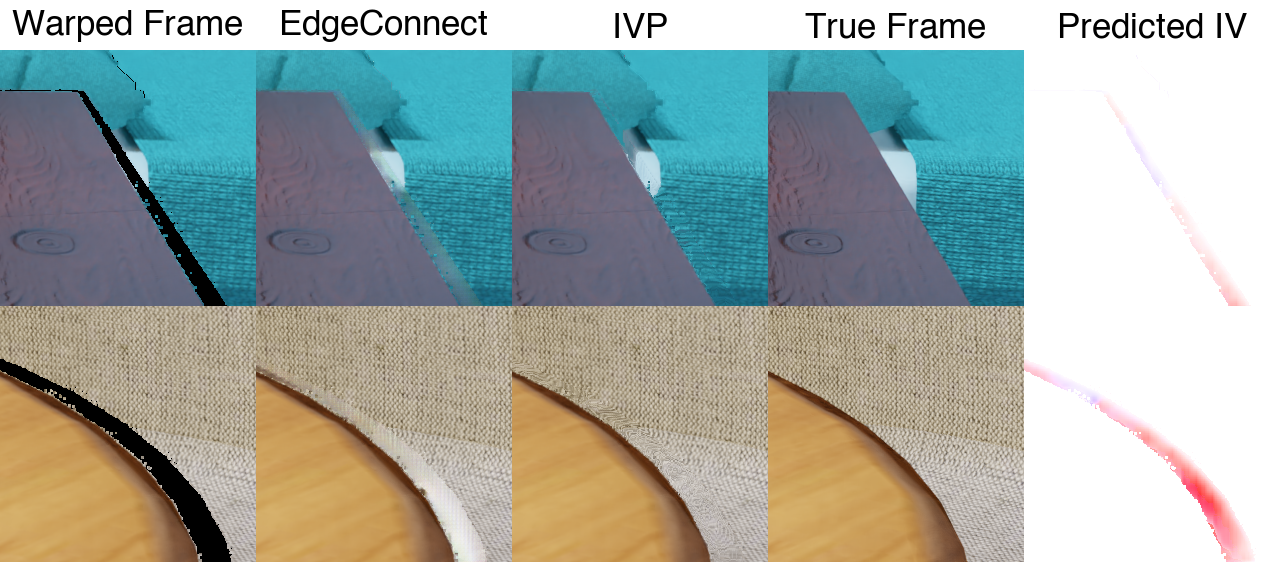}
        \caption{Texture and complex pattern reproduction by our model.}
        \label{fig:texture-reproduction}
    \end{figure}
    Nevertheless, we observe in Figure~\ref{fig:texture-reproduction} that our model can reproduce complex patterns and textures.
    We also show the infilling vectors predicted by our model in Figure~\ref{fig:texture-reproduction}, which we visualize similar to optical flow.
    We observe that the infilling vectors generated by our model are locally smooth.
    The smoothness of the infilling vectors allows a block of locations to be directly copied thereby preserving textures and patterns from the neighborhood.
     
     \begin{table}
        \scriptsize
        \centering
        \setlength\tabcolsep{1.75pt}
        \setlength{\extrarowheight}{2.75pt}
        \begin{tabular}{c|l|ccc|ccc}
            \hline
            \multirow{2}{*}{\makecell{Network\\Prediction}} & \multirow{2}{*}{TG \ \ Depth \ \ Iter\ \ } & \multicolumn{3}{c|}{IISc-VEED}&\multicolumn{3}{c}{SceneNet RGB-D}\\
            &&D-MSE &D-SSIM &TC&D-MSE&D-SSIM &TC\\
            \hline
            Intensity& \checkmark \qquad \checkmark  & 612 & 0.6956 & 158 & 1068 & 0.5633 & 266 \\[0.1cm]
            \hline
            IV & \quad \qquad \checkmark \qquad \checkmark & 464 & 0.7599 & 133 & 906 & 0.6137 & 241 \\
            IV & \checkmark \qquad \quad \qquad \checkmark & 457 & 0.7721 & 129 & 1049 & 0.5906 & 286 \\
            IV & \checkmark \qquad \checkmark & 473 & 0.7621 & 133 & 943 & 0.6068 & 254 \\
            IV & \checkmark \qquad \checkmark \qquad \checkmark & \textbf{442} & \textbf{0.7729} & \textbf{126} & \textbf{874} & \textbf{0.6230} & \textbf{240} \\
            \hline
        \end{tabular}
        \caption{Ablations. IV: Infilling Vector, TG: Temporal Guidance, Iter: Interative Infilling.}
        \label{tab:qa-ablations}
    \end{table}
    
    \textbf{Inference time} - The disocclusion infilling methods on average across scenes take between 2s to 5s per full HD frame on a single NVIDIA RTX 2080 Ti GPU, while the corresponding graphical rendering time in Blender is 150s. Nevertheless, infilling algorithms need to be optimized further, both in software and hardware, for real-time use cases.
    
    \subsection{Ablations}\label{subsec:ablations}
    
\begin{figure}
        \centering
        \includegraphics[width=0.95\linewidth]{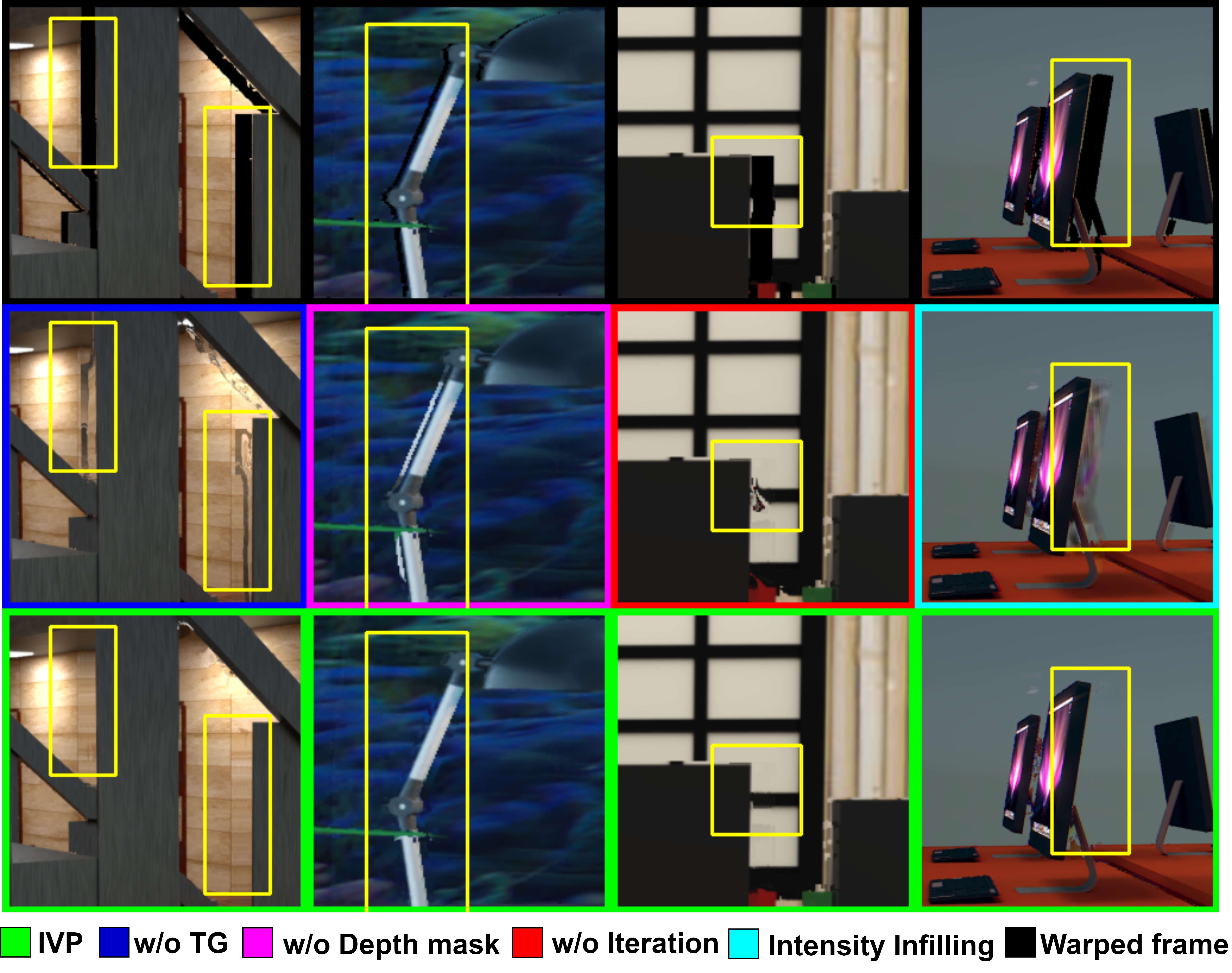}
        \caption{\textbf{Comparison of ablation models}.
        The first row shows warped frames, the second row shows predictions by different ablation models and the third row shows predictions by our complete model.
        See Section~\ref{subsec:ablations}.}
        \label{fig:ablations}
    \end{figure}

    \textbf{Infilling vector prediction vs. intensity prediction.}
    We compare our IVP approach to intensity prediction in the disoccluded regions, by using a U-Net based architecture that predicts the infilled frame directly.
    For fair comparison, we provide temporal information ($f_n$, $f^w_n$) and warped depth $d^w_{n+1}$ along with the warped frame $f^w_{n+1}$ as input to the network.
    From Table~\ref{tab:qa-ablations}, we see that infilling vector based models outperform the model that predicts the intensities directly.
    Figure~\ref{fig:ablations} shows that predicting the intensities directly leads to blurring artifacts where the network infills the disoccluded regions with the average intensity values of the boundary regions.
    However, our approach copies intensities from known regions, thereby leading to sharper predictions.
    
    \textbf{Importance of different components of our model.}
    We evaluate the benefit of temporal guidance, depth and iterative infilling by disabling one of them at a time.
    From Table~\ref{tab:qa-ablations}, we observe that temporal guidance is highly useful on IISc-VEED, where disocclusions are more correlated due to high frame rates.
   Since the SceneNet dataset has lower frame rate videos, we realize the importance of using ground truth depth.
    Nevertheless, on both datasets, using both the temporal prior and depth provides the best performance.
    We also observe improvement with iterative infilling which can remove and correct any errors that might have occurred in the initial iterations of infilling.

    In Figure~\ref{fig:ablations}, we observe that without temporal guidance, the model can find it difficult to read from the correct background object and end up copying from an incorrect object.
    We show an example where the camera trajectory changes and the lack of use of depth can lead to poorer performance.
    Finally, we also observe that iterative infilling can correct erroneous infilling in previous iterations.

    \section{Conclusion}\label{sec:conclusion}
    We introduced a learning based infilling vector prediction model for revealing disocclusions in temporal view synthesis. We showed how guiding the infilling vector prediction using a temporal prior and depth can better exploit the structure of the problem. We introduced a temporal consistency measure to show that our predictions are temporally consistent. Since we observe that warping followed by infilling typically consumes much less time compared to graphically rendering video frames, temporal view synthesis appears to be a promising approach to increase frame rates in low compute VR devices.\\
    \textbf{Acknowledgments} This work was supported by a grant from Qualcomm. We also thank Vinay Melkote and Ajit Rao for helpful discussions and comments.

        {\small
    \bibliographystyle{ieee_fullname}
    \bibliography{VSTU}
    }

\end{document}